\documentclass{article}
\usepackage{ijcai22}

\usepackage{times}
\usepackage{soul}
\usepackage{url}
\usepackage[hidelinks]{hyperref}
\usepackage[utf8]{inputenc}
\usepackage[small]{caption}
\usepackage{graphicx}
\usepackage{amsmath}
\usepackage{amsthm}
\usepackage{booktabs}
\usepackage{algorithm}
\usepackage{algorithmic}
\usepackage{multirow}
\usepackage{subfigure}
\usepackage{amsfonts}
\title{ViT2Hash: Unsupervised Information-Preserving Hashing}

\author{
Qinkang Gong$^1$\and
Liangdao Wang$^1$\and
Hanjiang Lai$^1$\footnote{Contact Authors}\and
Yan Pan$^1$\And
Jian Yin$^{1\ast}$\\
\affiliations
$^1$Sun Yat-Sen University\\
\emails
\{gongqk, wangld5\}@mail2.sysu.edu.cn,
\{laihanj3, panyan5, issjyin\}@mail.sysu.edu.cn
}

\begin{document}

\maketitle

\begin{abstract}
Unsupervised image hashing, which maps images into binary codes without supervision, is a compressor with a high compression rate. Hence, how to preserving meaningful information of the original data is a critical problem. Inspired by the large-scale vision pre-training model, known as ViT, which has shown significant progress for learning visual representations, in this paper, we propose a simple information-preserving compressor to finetune the ViT model for the target unsupervised hashing task. Specifically, from pixels to continuous features, we first propose a feature-preserving module, using the corrupted image as input to reconstruct the original feature from the pre-trained ViT model and the complete image, so that the feature extractor can focus on preserving the meaningful information of original data. Secondly, from continuous features to hash codes, we propose a hashing-preserving module, which aims to keep the semantic information from the pre-trained ViT model by using the proposed Kullback-Leibler divergence loss. Besides, the quantization loss and the similarity loss are added to minimize the quantization error. Our method is very simple and achieves a significantly higher degree of MAP on three benchmark image datasets. 
\end{abstract}

\section{Introduction}


Due to the fast retrieval speed and low storage cost of binary codes, deep learning-to-hash~\cite{yuan2020central} has attracted extensive attention in the past decades and has been successfully applied to large-scale image retrieval. 
Since collecting large numbers of manual annotations is time-consuming and labor-intensive, much research effort has been paid to the unsupervised hashing methods~\cite{shen2020auto,zhang2020deep}, which learn the similarity-preserving binary codes without annotations. 

To solve the problem of no labels, the existing unsupervised hashing methods can be roughly divided into two categories~\cite{DBLP:conf/ijcai/QiuSOYC21}: reconstruction-based methods and non-reconstruction-based methods. Reconstruction-based methods~\cite{dizaji2018unsupervised,shen2020auto} try to reconstruct the original input image. For example, generative adversarial network~\cite{goodfellow2014generative} is applied to unsupervised hashing~\cite{zieba2018bingan,dizaji2018unsupervised}. The non-reconstruction-based methods use extra information to learn the similarity-preserving binary codes, \textit{e.g.}, contrastive learning~\cite{lin2016learning,DBLP:conf/ijcai/QiuSOYC21}, graphs~\cite{shen2018unsupervised,shen2020auto} and pseudo labels~\cite{hu2017pseudo,su2018greedy}. 

\begin{figure}[t]
\centering
\includegraphics[width=0.9\linewidth]{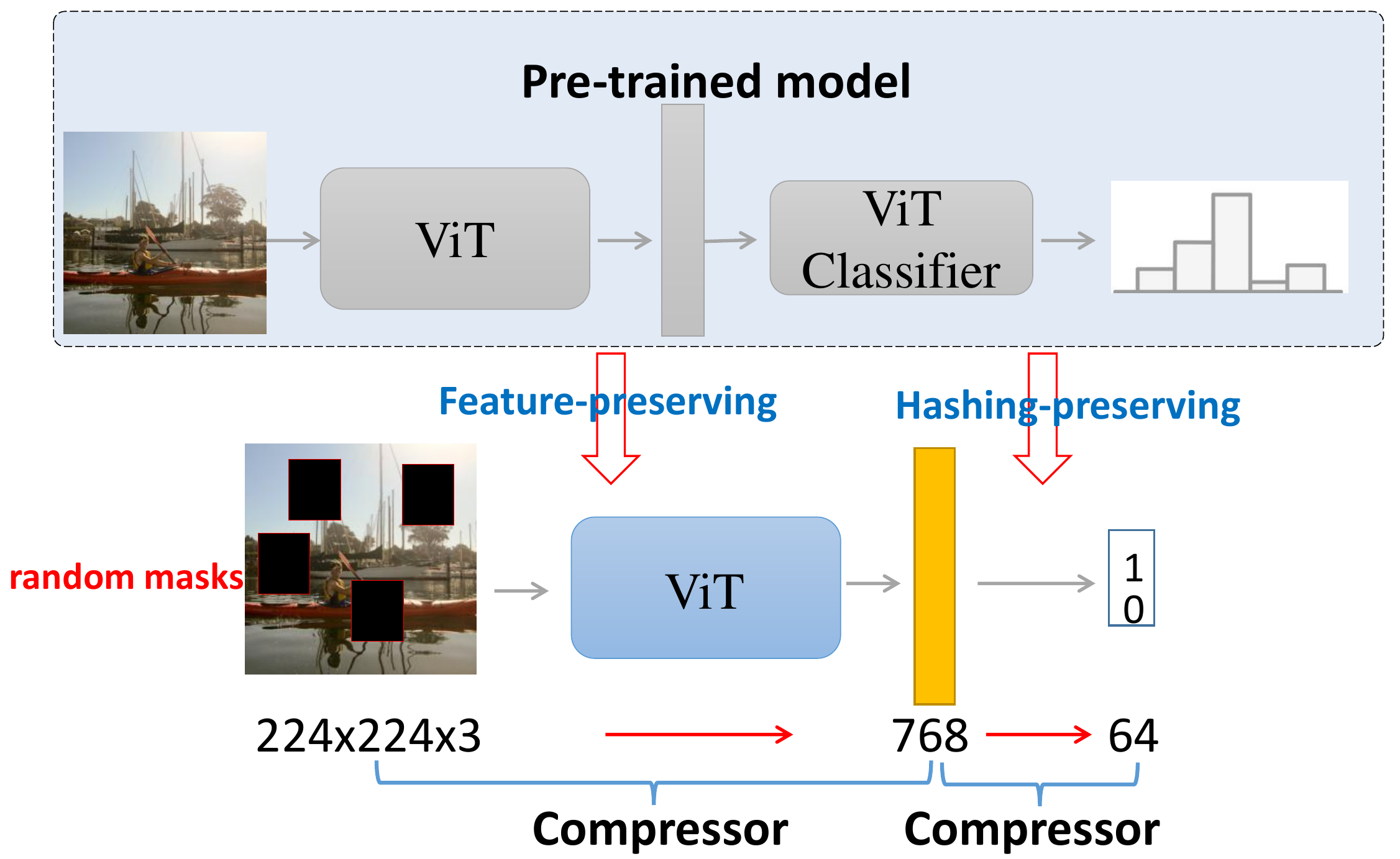}
\caption{Illustration of the proposed method. We employ to finetune the pre-trained ViT model  for the hashing task. The hashing task is a compressor from pixels to continuous features to binary codes, which will cause huge information loss. In this paper, we propose information-preserving modules to preserve the meaningful information.  
}
\label{information_loss}
\end{figure}

Recently, using the pre-trained models that pretraining on large-scale datasets becomes the de-facto standard for deep learning. The pre-trained model, known as Vision Transformer (ViT)~\cite{dosovitskiy2020image}, attains inspirational performances on various tasks, which is based on the Transformer~\cite{vaswani2017attention} architecture and achieves significant progress in visual representations. 
However, most previous unsupervised hashing methods using pre-trained models do not fully explore how to preserve meaningful information with such a high compression rate. As shown in Figure~\ref{information_loss}, an input image usually contains lots of pixels (\textit{e.g.}, $224 \times 224 \times 3$), which will be encoded to $768$-d continues feature and finally compressed to $64$-bit binary code. As the dimension of the data keeps getting smaller, the semantic information it contains will continue losing inevitably. Therefore, it is beneficial to consider how to reduce information loss in these two stages.

To address the above problem, we propose a simple unsupervised Information-Preserving Hashing (IPHash) to preserve the semantic information and improve the retrieval ability of hash codes as much as possible. First, from pixels to continuous features, 
inspired by MAE~\cite{he2021masked}, we can know that the image has heavy spatial redundancy, and the complete image information can be reconstructed only by random masking part of the image. 
Observed by this, we use the random masked images as inputs and restrict the output features to be consistent with the pre-trained ViT features extracted from the original images. Compared with the original pre-trained ViT model, the new feature extractor can better preserve the meaning information since it focuses on preserving and reconstructing the information of original data. 


Secondly, from continuous features to hash codes, we 
use knowledge distillation~\cite{hinton2015distilling} to transfer the semantic information from the pre-trained ViT model to the hash codes. The semantic information of the ViT Classifier is used as the extra information and the Kullback-Leibler divergence loss is proposed to minimize distance between the classifier of binary codes and the pre-trained ViT classifier.  
Therefore, the hash codes can fully preserve the semantic information of the pre-trained model. Besides, the similarities between the binary codes and the continuous pre-trained features~\cite{kulis2009learning} as well as the quantization error are also considered, which are widely used loss objectives in unsupervised hashing~\cite{su2018greedy}.  


In summary, our contributions are listed as follows:

\begin{itemize}
\item In this work, we propose a novel and simple unsupervised information-preserving hashing algorithm, which can effectively fine-tune pre-trained ViT model for the hashing task.   

\item The two information-preserving modules are proposed for effectively reducing the information loss from pixels to continuous features and then to binary codes, so as to improve the retrieval performance.
\item We evaluate our proposed model on three benchmark datasets CIFAR-10, MSCOCO and NUS-WIDE and it significantly outperforms the previous baselines.
\end{itemize}

\section{Related work}

\paragraph{Unsupervised hashing methods} 
The traditional unsupervised hashing methods use hand-crafted features and shallow hash functions to obtain hash codes. For instance, spectral hashing~\cite{weiss2008spectral} generates efficient binary codes by spectral graph partitioning. ITQ~\cite{gong2012iterative} minimizes the quantization error between PCA-projected data and binary codes by finding the proper rotation matrix. Due to the limitation of hand-crafted features, deep neural networks are introduced to extract better hash codes. The existing deep unsupervised hashing methods can be roughly divided into reconstruction-based methods and non-reconstruction-based methods. 

Reconstruction-based methods try to reconstruct the original image using the hash codes. 
TBH~\cite{shen2020auto} is an auto-encoder with two kinds of bottlenecks to learn optimal binary codes. Some methods also apply the generative adversarial network. For example, HashGAN~\cite{dizaji2018unsupervised} enforces the synthesized image and the real image have the same binary hash code. 

On the other hand, there are also some non-reconstruction-based methods. For instance, Deep Hashing (DH)~\cite{erin2015deep} adds three constraints on the top layer of the deep network: quantization loss, balanced bits, and independent bits. The most widely used extra information is the pre-trained model, \textit{e.g.}, \cite{hu2017pseudo} clustered images via K-means using the extracted features from the pre-trained model and treated them as pseudo labels. In \cite{su2018greedy}, the cosine similarity between any two hash codes should be consistent with their corresponding continuous features extracted by deep pre-trained models. Recently, \cite{DBLP:conf/ijcai/QiuSOYC21} proposed a contrastive learning method for unsupervised hashing. 

\paragraph{Vision Transformer}
Inspired by the success of Transformer~\cite{vaswani2017attention} in NLP tasks, Transformer architecture are widely applied in computer vision tasks~\cite{carion2020end,arnab2021vivit}. Most recently, Vision Transformer (ViT)~\cite{dosovitskiy2020image} regularly partitions the image into 16x16 patches, which will be fed into the pure Transformer architecture as a sequence and it is the first work to show that a pure Transformer based on image classification model can achieve the state-of-the-art. 

\paragraph{Masked image encoding}
Inspired by the mask language model in NLP~\cite{devlin2019bert}, many recent works based on Transformer~\cite{vaswani2017attention} try to explore the utilization of masking in image task, \textit{e.g.}, BEiT~\cite{bao2021beit} proposed to predict masked discrete tokens. Recently, Masked Auto-Encoder (MAE)~\cite{he2021masked} is proposed based on randomly masking some patches in an image and trying to reconstruct them using self-supervised learning. MAE shows that images are highly redundant representations of natural information that a missing patch can be recovered by its neighboring patches. Therefore, randomly masking a high portion of patches can reduce information redundancy and preserve more compact semantic information. 


\begin{figure*}[t]
\centering
\includegraphics[width=\linewidth]{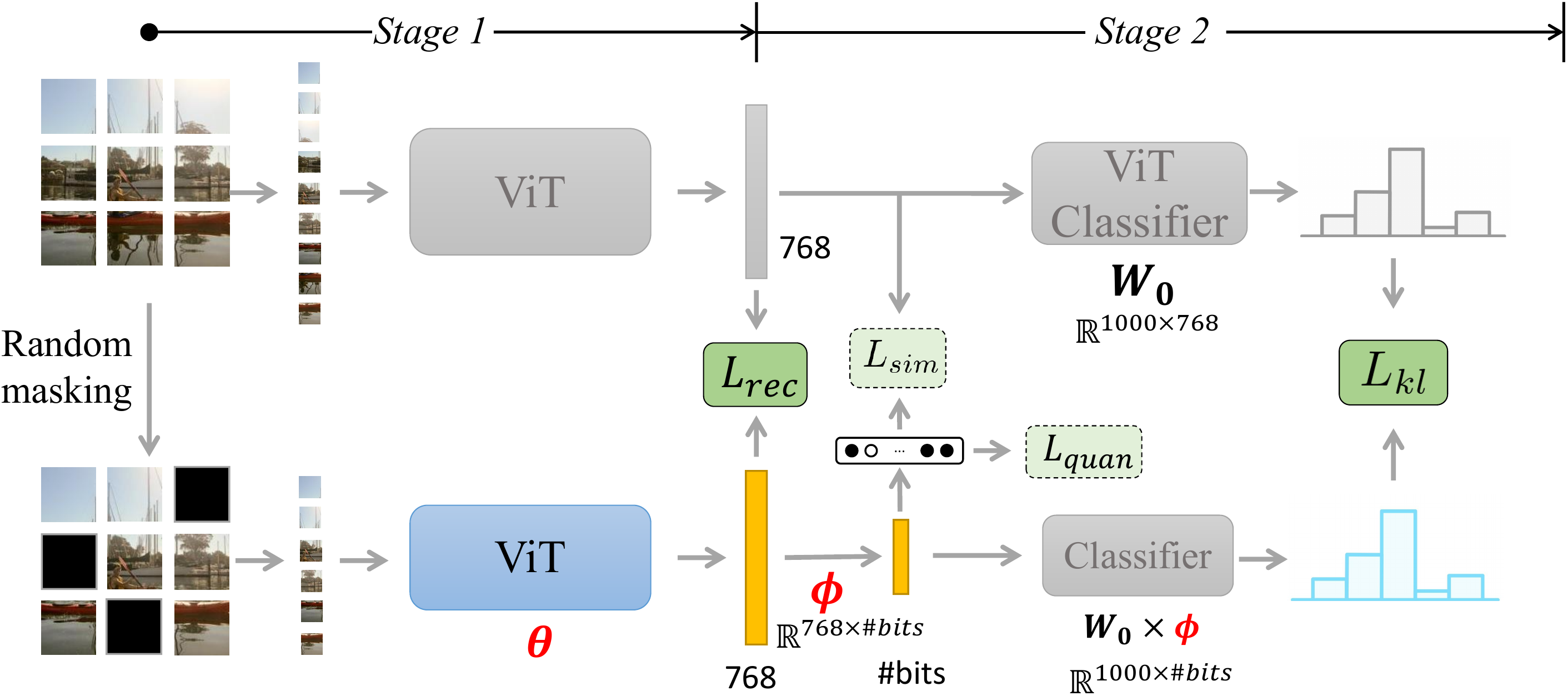}
\caption{The pipeline of our proposed unsupervised Information-Preserving Hashing (IPHash). It consists of two stages and only two parameters are needed to be learnt, (Stage 1) from image pixel data to floating-point continuous features (feature extractor $\theta$), (Stage 2) from floating-point continuous features to binary codes (a hashing layer $\phi$). Four information-preserving losses are proposed to finetune the pre-trained ViT model.}
\label{IPHash_model}
\end{figure*}

\section{Unsupervised Information-Preserving Hashing}
In this section, we first give the problem definition. Given the unlabeled training set $X=\{x_i\}_{i=1}^N$, the goal of unsupervised hashing is to learn a mapping function that encodes input images into binary codes, and the similarities among these codes are preserved. The main problem of unsupervised hashing is how to preserve the similarities without supervision. 

Since the “pretraining-finetuning” paradigm becomes the de-facto standard for deep learning,  in this paper, we propose an \textbf{I}nformation-\textbf{P}reserving \textbf{Hash}ing (IPHash), which is an unsupervised hashing approach and finetunes the pre-trained model ViT to preserve the information without any supervision. 
Formally, we divide the hashing into two stages: from image pixels to continuous features, and from continuous features to hash codes. 
As shown in Figure~\ref{IPHash_model}, IPHash consists of two modules: feature-preserving module and hashing-preserving module.  In the following, we will introduce details of our proposed method. 


\subsection{Feature-Preserving Module}

Recently, masked autoencoder (MAE~\cite{he2021masked})  has been received considerable attention in self-supervised pre-training task. It shows that the image has heavy spatial redundancy. As we can see that hashing is a compressor with very high compression rate, thus it is important that the feature extractor of hashing is better to preserve the meaningful information and throw away the redundancy. Inspired by MAE, we also use the subset of non-overlapping patches to train the feature extractor
\begin{equation}
\label{RM}
\tilde{x}_i=RM(x_i;m),
\end{equation}
where $RM$ denotes random masking operation and $m$ denotes the masking ratio. 

From pixels to continuous features, the feature extractor $f(\cdot;\theta)$ is utilized to extract $d$-dimensional continuous features $V=\{v_1, v_2,...,v_N|v_i\in\mathbb{R}^{d}\}$ as following
\begin{equation}
\label{continuous_features}
v_i=f(\tilde{x}_i;\theta).
\end{equation}
In this paper, we use the ViT as our backbone and the output of class token as continuous features $v_i$. The parameter $\theta$ is initialized by the pre-trained model ViT, and we need to fine-tune $\theta$ to achieve better feature representations for hash codes. 

 In this paper, we propose to use the fixed pre-trained ViT model to provide the pseudo annotations for keeping semantic information. Formally, the $d$-dimensional pseudo annotation for $i$-th sample's feature is defined as 
\begin{equation}
\label{continuous_features_ViT}
v_{i}^{\text{ViT}}=f(x_i;\text{ViT}).
\end{equation}
where $v_{i}^{\text{ViT}}$ is the extracted feature from the completed image and is used as the pseudo annotation.

As shown in stage 1 in Figure \ref{IPHash_model}, we calculate the reconstruction loss between the features of corrupted images and the target features as following
\begin{equation}
\label{L_rec}
L_{rec}=\frac{1}{N}\sum_{i=1}^{N}\Vert f(\tilde{x}_i;\theta) - v_{i}^{\text{ViT}} \Vert^2_2.
\end{equation}
In later experiments, the feature extractor fine-tuned with the mask images performs better than the original pre-trained model. This is because the feature extractor is learned to better preserve and reconstruct the meaningful information of original data. 

\subsection{Hashing-Preserving Module}

Now we have feature $v_i$ from the masked image, then we use a simple linear layer to encode  the continuous features into low dimensional feature
\begin{equation}
\label{h_codes}
h_i= v_i  \times  \phi,
\end{equation}
where $\phi \in \mathbb{R}^{768 \times \#bits} $ is a projection matrix that needed to be learned and $\#bits$ is the desired length of binary codes. Since there is no hash layer in the pre-trained ViT model, we use the random initialization of $\phi$. The hash code $b_i$ is obtained after binarizing $h_i$. 

Similar in the first stage, we try to preserve the information from the pre-trained ViT classifier. The output of ViT classifier with the complete image as input is used as the pseudo soft label, which is 
\begin{equation}
\label{gt_logits}
z_i^{\text{ViT}}= \text{Softmax}(W_0 \times v_{i}^{\text{ViT}}   / \tau),
\end{equation}
where the above is the pre-trained ViT classifier and $W_0 \in \mathbb{R}^{1000 \times 768}$ is the original pre-trained parameters of ViT classifier. The softmax function uses the temperature $\tau$.

While the hashing task does not have the classifier. To fully utilize the knowledge from the pre-trained ViT model, we also construct a classifier after the hash layer, which is formulated as
\begin{equation}
\label{hash_logits}
z_i= \text{Softmax}( W_0 \times \phi \times h_i / \tau),
\end{equation}
where $W = W_0 \times \phi$ is the constructed classifier and $W_0$ is fixed. Please note that this classifier uses the $ W_0 \times \phi$ as the matrix projection. We use such parameters to constrain the range of the hashing layer $\phi$ to preserve the information from the pre-trained ViT classifier.   


The objective is defined as
\begin{equation}
\label{L_kl}
L_{kl}=\frac{1}{N}\sum_{i=1}^{N}KL(z_i, z_i^{\text{ViT}} ),
\end{equation}
where $KL$ denotes the Kullback-Leibler divergence loss. 

To explain why we use $z_i^{\text{ViT}}$ as the pseudo soft label, the predicted probabilities by the pre-trained classifier contain the similarity information between the samples, \textit{e.g.}, the smaller of KL distance between two predicted probabilities, the more similar between these two images. Thus we can use them as the extra information to learn the better hash codes. It can also be viewed as the knowledge distillation from the ViT model to the downstream hashing task. 


Besides, to better reduce the quantization error, we also use the $v_{i}^{\text{ViT}}$ as the extra information to preserve the similarities between the continue features and the hash codes~\cite{kulis2009learning}, which is defined as
\begin{equation}
\label{sim_loss}
L_{sim}=\frac{1}{M}\sum_{i,j}\Vert \cos(v_{i}^{\text{ViT}},v_{j}^{\text{ViT}})-\cos(b_i,b_j) \Vert^2_2,
\end{equation}
where $cos(\cdot,\cdot)$ denotes cosine similarity and $M$ denotes the number of pairs. Following the design of \cite{su2018greedy}, the optimization of sign function is done with straight-through estimator~\cite{bengio2013estimating} which the gradients are transmitted intactly to the front layer in back propagation.

The quantization loss also should be minimized, which is defined as
\begin{equation}
\label{quan_loss}
L_{quan}=\frac{1}{N}\sum_{i=1}^N\Vert h_i-b_i\Vert_2^2=\frac{1}{N}\sum_{i=1}^N\Vert h_i-sign(h_i)\Vert_2^2,
\end{equation}

\subsection{Training objectives}
Finally, IPHash is trained by jointly minimizing all the above losses as follows
\begin{equation}
\label{L_final}
L=L_{kl} + L_{sim} + L_{quan} + \gamma  L_{rec},
\end{equation}
where $\gamma$ aims to balance the weight of two stages. In this paper, since the feature extractor in the first stage is initialized by the pre-trained ViT model, we use the smaller weight, \textit{i.e.}, $\gamma=0.1$, in our all experiments. 

\subsection{Testing}
\begin{figure}[h]
\centering
\includegraphics[width=0.9\linewidth]{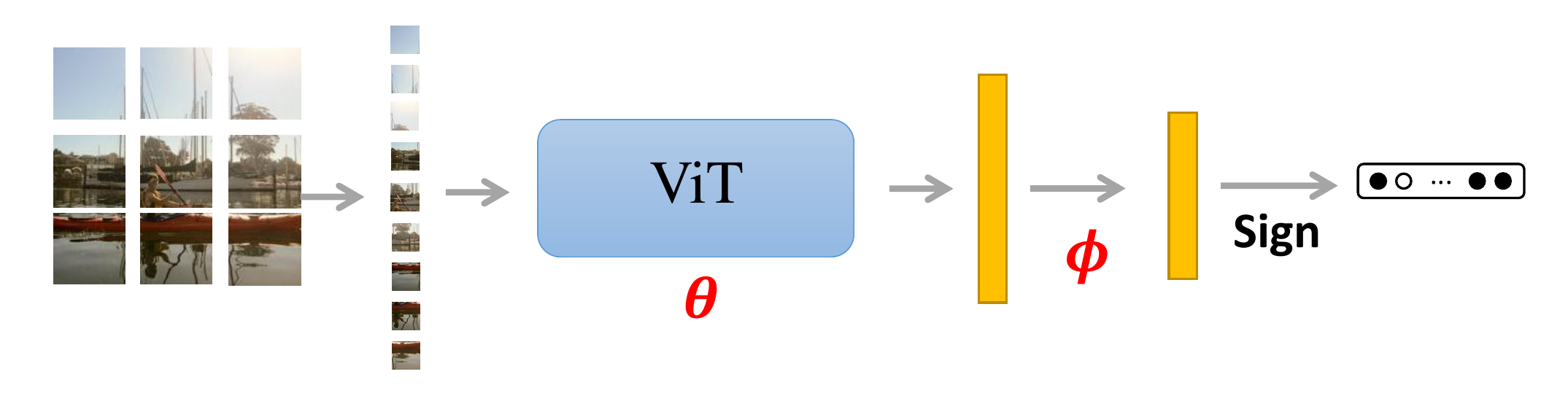}
\caption{Illustration of the proposed method in testing. 
}
\label{testing}
\end{figure}

After training the feature extractor $\theta$, the fully-connected hash layer $\phi$, we use the complete images as inputs to encode images into binary codes as shown in Figure~\ref{testing}.

\begin{table*}[!t]
\centering
\caption{MAP w.r.t different numbers of bits.}
\label{result_all_datasets}
\begin{tabular}{c|c|c c c|c c c|c c c}
\hline
    \multirow{2}{*}{Methods} & \multirow{2}{*}{Backbone} & \multicolumn{3}{c|}{CIFAR-10} & \multicolumn{3}{c|}{NUS-WIDE} & \multicolumn{3}{c}{MSCOCO}  \\
    ~ & ~ & 16bits & 32bits & 64bits & 16bits & 32bits & 64bits & 16bits & 32bits & 64bits \\
\hline
\hline
    ITQ & VGG & 0.305 & 0.325 & 0.349 & 0.627 & 0.645 & 0.664 & 0.598 & 0.624 & 0.648 \\
    DeepBit & VGG & 0.194 & 0.249 & 0.277 & 0.392 & 0.403 & 0.429 & 0.407 & 0.419 & 0.430 \\
    BinGAN & - & 0.476 & 0.512 & 0.520 & 0.654 & 0.709 & 0.713 & 0.651 & 0.673 & 0.696 \\
    GreedyHash & VGG & 0.448 & 0.473 & 0.501 & 0.633 & 0.691 & 0.731 & 0.582 & 0.668 & 0.710 \\
    HashGAN & - & 0.447 & 0.463 & 0.481 & - & - & - & - & - & - \\
    DVB & VGG & 0.403 & 0.422 & 0.446 & 0.604 & 0.632 & 0.665 & 0.570 & 0.629 & 0.623 \\
    TBH & VGG & 0.532 & 0.573 & 0.578 & 0.717 & 0.725 & 0.735 & 0.706 & 0.735 & 0.722 \\
    CIBHash & VGG & 0.590 & 0.622 & 0.641 & 0.790 & 0.807 & 0.815 & 0.737 & 0.760 & 0.775 \\
\hline
    ITQ & ViT & 0.870  & 0.901  & 0.910  & 0.724  & 0.756  & 0.779 & 0.715 & 0.805 & 0.844 \\
    GreedyHash & ViT & 0.879 & 0.901 & 0.915 & 0.629 & 0.690 & 0.752 & 0.647 & 0.756 & 0.836 \\
    CIBHash & ViT & 0.903 & 0.925 & 0.938 & 0.779 & 0.810 & \textbf{0.826} & 0.809 & 0.846 & 0.867 \\
\hline
    IPHash(ours) & ViT & \textbf{0.942} & \textbf{0.951} & \textbf{0.958} & \textbf{0.797} & \textbf{0.816} & \textbf{0.826} & \textbf{0.826} & \textbf{0.875} & \textbf{0.894}  \\
\hline
\end{tabular}
\end{table*}

\section{Experiments}
                    
\subsection{Experimental settings}

\paragraph{Datasets.} Three popular benchmark datasets: CIFAR-10, MSCOCO, and NUS-WIDE are used to evaluate the proposed hashing method. \textbf{CIFAR-10}~\cite{krizhevsky2009learning} consists of 60,000 images from 10 classes. We follow the settings in~\cite{DBLP:conf/ijcai/QiuSOYC21} and randomly select 1,000 images per class as the query set and the remaining images as the database. Then we uniformly select 5,000 unlabeled images from the database as the training set. \textbf{MSCOCO}~\cite{lin2014microsoft} consists of 123,287 images from 80 classes in total. We randomly select 5,000 images as the query set and the remaining data are utilized as the retrieval set. 10,000 images sampled from the retrieval set are selected as the unlabelled training set. \textbf{NUS-WIDE}~\cite{chua2009nus} consists of 269,468 images of 81 categories. Following the settings in \cite{xia2014supervised}, we adopt the subset of images from the 21 most frequent categories. We randomly select 2,100 images as a query set and employ 10,500 unlabeled images for training.

\paragraph{Compared Methods.} We compared our methods with various unsupervised hashing methods including \textbf{ITQ}~\cite{gong2012iterative}, \textbf{GreedyHash}~\cite{su2018greedy} and \textbf{CIBHash}~\cite{DBLP:conf/ijcai/QiuSOYC21}. For fair comparisons, we replaced the backbone of the these three baseline algorithms with ViT-16/B, the same with our method. In addition, we also compare with previous methods based on VGG-16 or reconstruction architecture, including \textbf{DeepBit}~\cite{lin2016learning}, \textbf{BinGAN}~\cite{zieba2018bingan}, \textbf{HashGAN}~\cite{dizaji2018unsupervised}, \textbf{DVB}~\cite{shen2019unsupervised}, \textbf{TBH}~\cite{shen2020auto}. For the reported performance of baselines, they are quoted from CIBHash~\cite{DBLP:conf/ijcai/QiuSOYC21}.

\paragraph{Implementation details.} Our method is implemented using PyTorch framework. We take ViT-B/16~\cite{dosovitskiy2020image} as the backbone network in our experiments. In all experiments, we use Adam as optimizer with the initial learning rate of $1 \times 10^{-3}$ for hashing layer (the parameters are randomly initialized and thus use the larger learning rate) and $1 \times 10^{-5}$ for feature extractor (it is initialized with the pre-trained ViT model and the smaller learning rate is applied for finetuning). We run the models in 100 epochs and the batch size is set to 64 for all experiments. The default mask ratio $m$ is set to 25\%.

\paragraph{Evaluation Metrics.} In this paper, we use Mean Average Precision (MAP) as our evaluation metric. The MAP is used to measure the accuracy of the whole binary codes based on the Hamming distances. It is computed as the mean of average precision(AP) for the whole queries:
\begin{equation}
\label{AP}
AP=\frac{1}{n_{+}}\sum_{k=1}^{n}P_{k}\times rel_{k},
\end{equation}
where $n_{+}$ is the number of relevant samples in the ranking list, $n$ is the size of retrieval database, $P_{k}$ is the precision in the top-$k$ returns. And $rel_{k}=1$ when the $k$-th returned sample in the ranking list is relevant,  otherwise $rel_{k}=0$. We follow the settings in \cite{cao2017hashnet} and use MAP@1000 for CIFAR-10, MAP@5000 for MSCOCO and NUS-WIDE.

\subsection{Experimental results}

\paragraph{Overall Performance.} The details of the comparative results on three datasets are presented in Table~\ref{result_all_datasets}. First of all, since the power of the feature representations of ViT pre-trained model, approaches using the ViT backbone perform better than the previous methods.   
Secondly, we can observe that our method also performs best for all ViT-based methods. For example, compared with the recent state-of-the-art method CIBHash, our method brings an averaged increase of 2.8\%, 0.8\%, 2.4\% (averaged over different code lengths) on CIFAR-10, NUS-WIDE, and MSCOCO datasets, respectively. This reveals that our IPHash can better preserve semantic information in the hash codes. 

\begin{figure*}[t]
\subfigure[16 bits]{
\begin{minipage}[t]{0.3\linewidth}
\centering
\includegraphics[width=\linewidth]{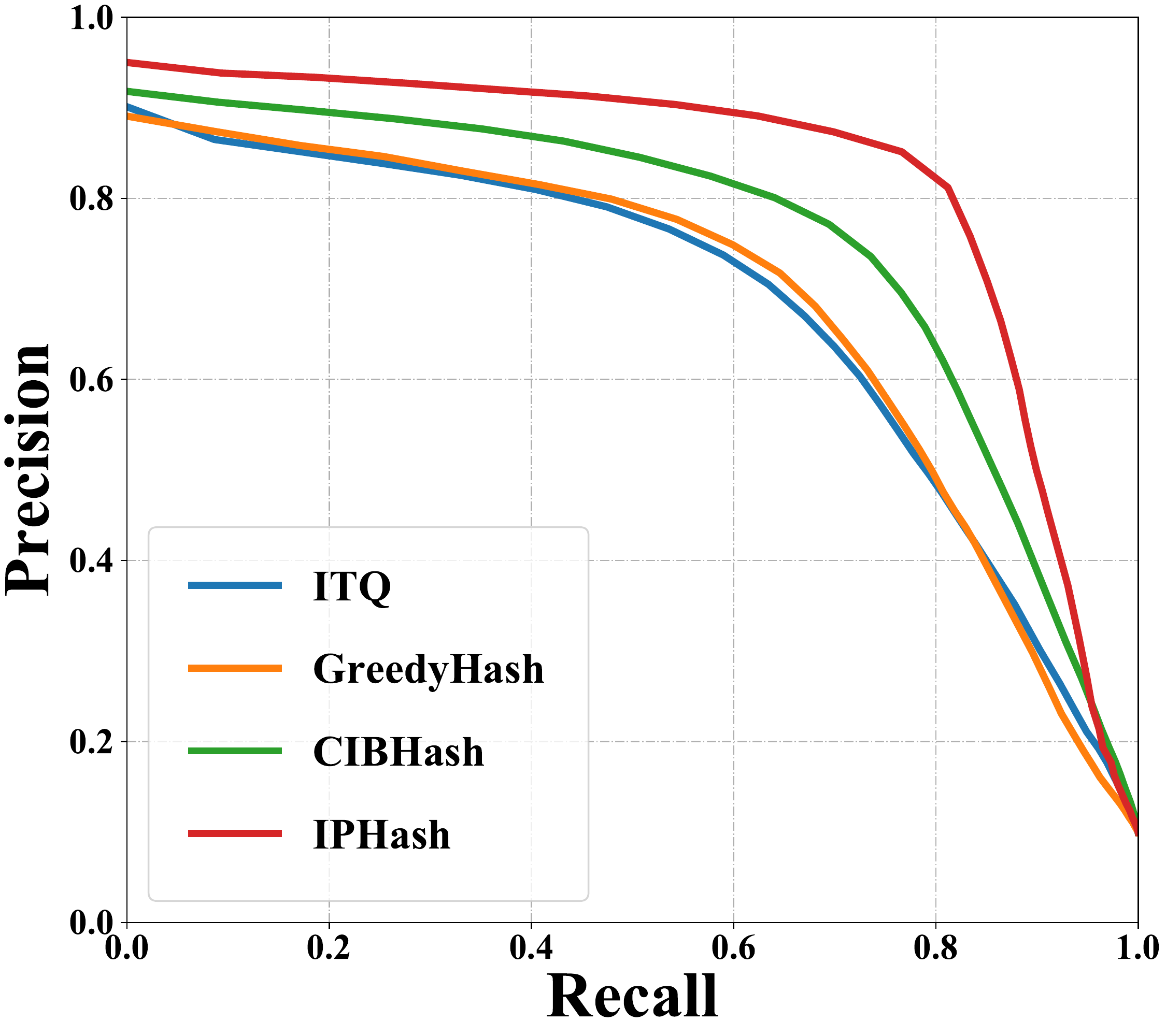}
\end{minipage}
}%
\subfigure[32 bits]{
\begin{minipage}[t]{0.3\linewidth}
\centering
\includegraphics[width=\linewidth]{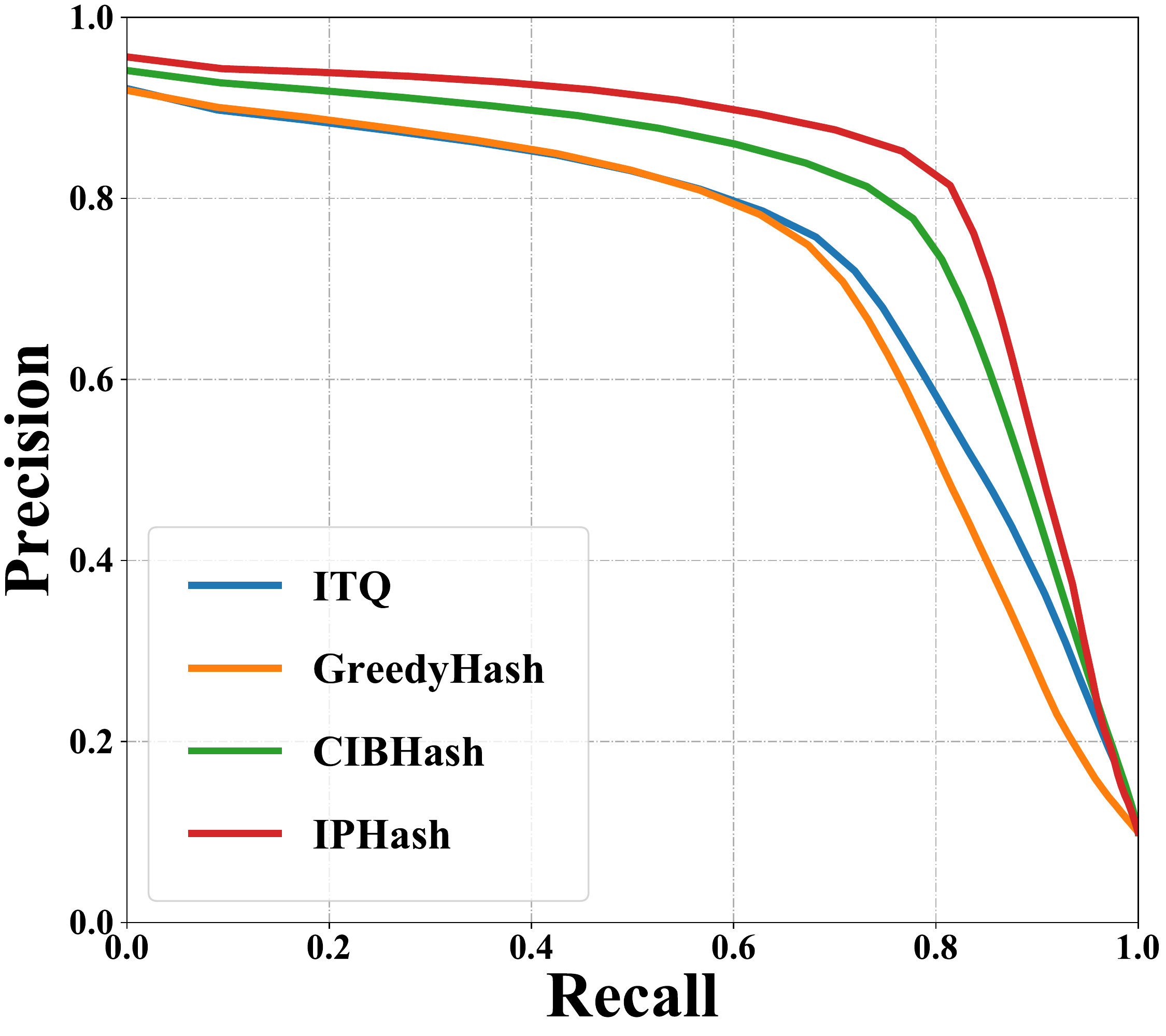}
\end{minipage}
}%
\subfigure[64 bits]{
\begin{minipage}[t]{0.3\linewidth}
\centering
\includegraphics[width=\linewidth]{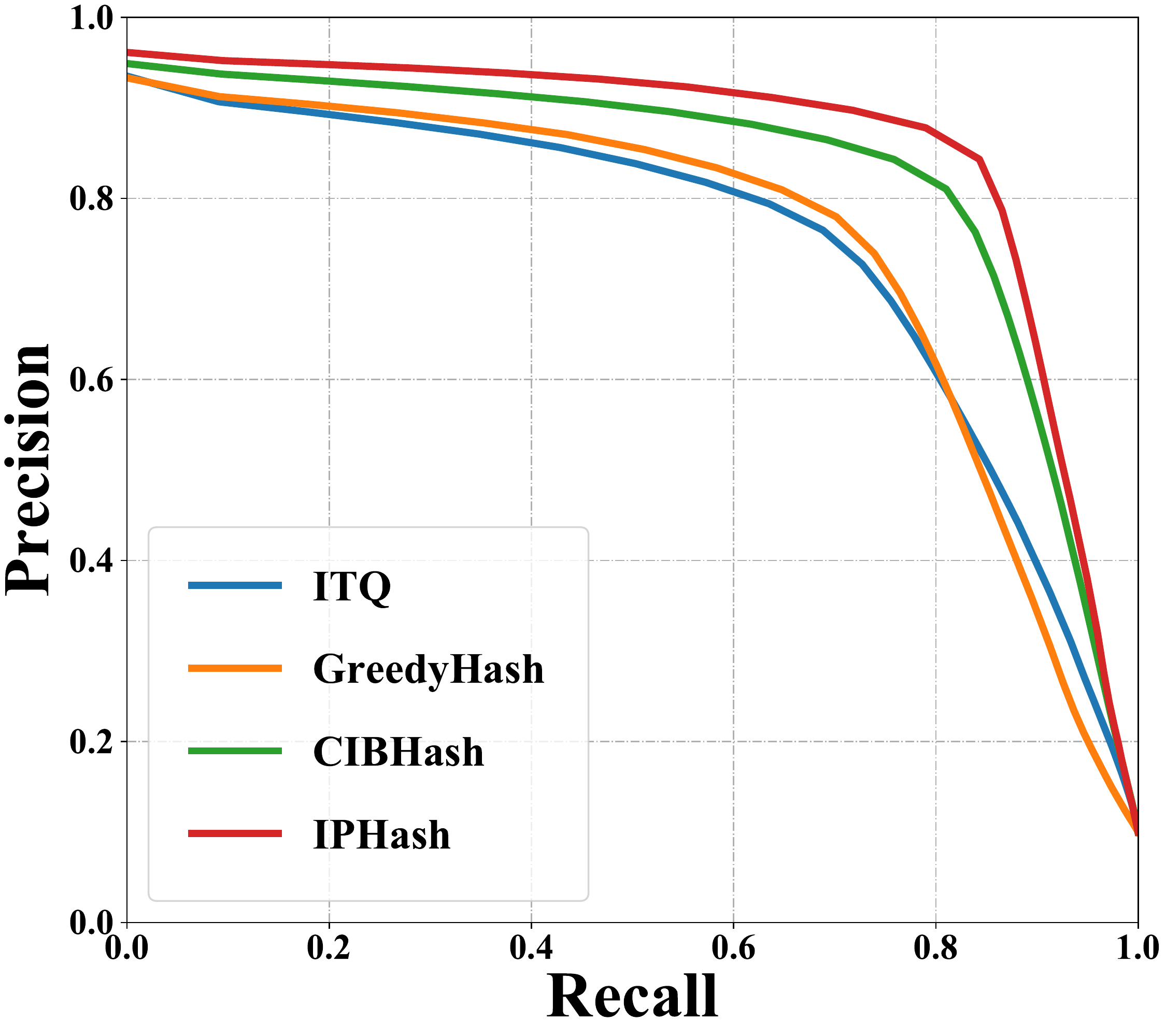}
\end{minipage}
}%
\centering
\caption{Precision-Recall curves on CIFAR-10.}
\label{pr_curve_on_cifar}
\end{figure*}

\paragraph{Precision-Recall Curve.} Figure \ref{pr_curve_on_cifar} shows the Precision-Recall (P-R) curve on CIFAR-10 of different methods based on ViT. As can be seen from the result, our retrieval performance is obviously better than CIBHash, GreedyHash, and ITQ, especially at 16-bit. The reason may be that when the code length is shorted, it has a higher compression rate. The proposed method can help to keep the meaningful information at such a higher compression rate. 

\begin{figure}[h]
\centering
\includegraphics[width=\linewidth]{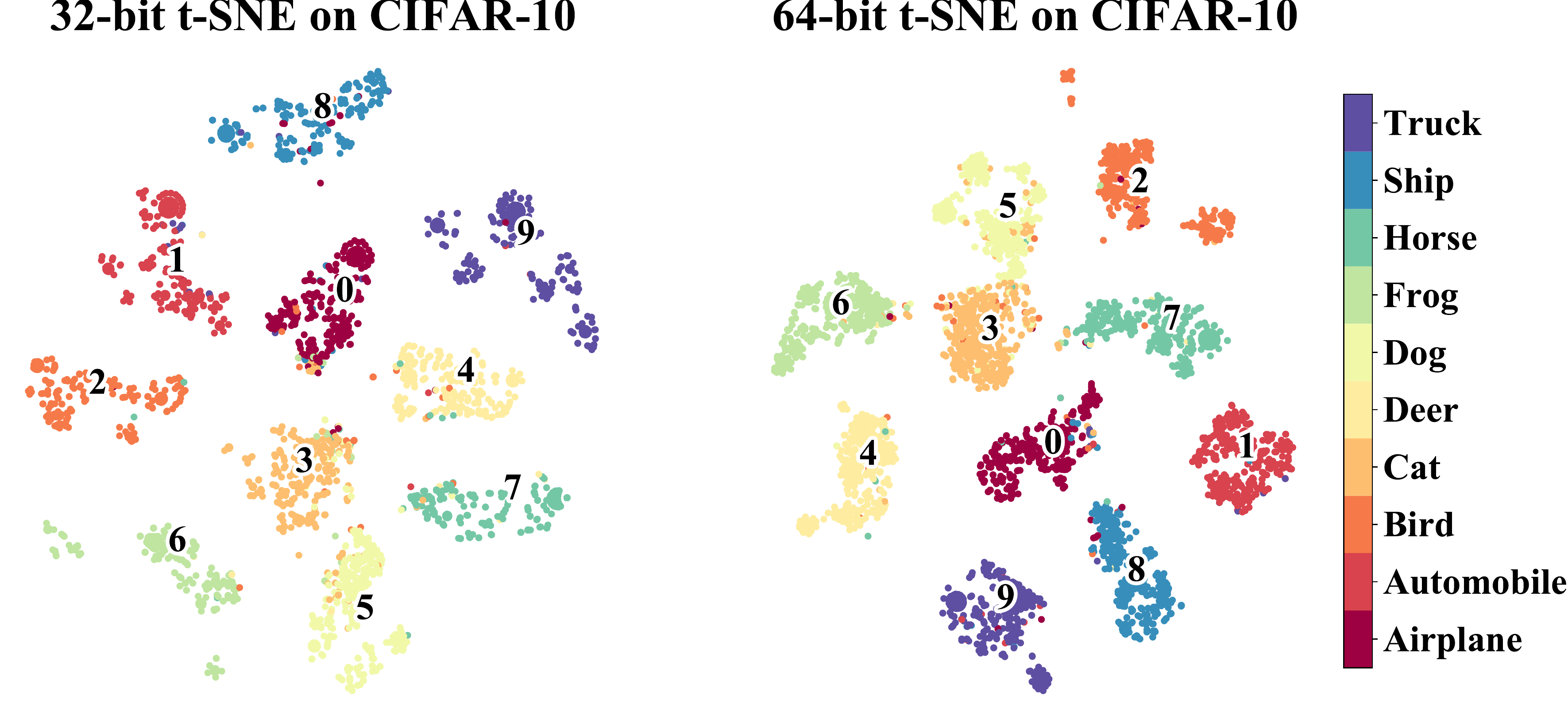}
\caption{T-SNE visualization on CIFAR-10.}
\label{visualization_on_cifar}
\end{figure}

\paragraph{Visualization.} Figure \ref{visualization_on_cifar} shows the T-SNE visualization results of IPHash on CIFAR-10. The samples of different semantics (labels) are clearly distinguished both at 32-bit and 64-bit, which shows the powerful information-preserving ability of our model.

\begin{table}
\centering
\caption{MAP comparison with variants of IPHash.}
\label{ablation_study}
\begin{tabular}{c|c|ccc}
\hline
\multicolumn{2}{c|}{\textbf{Component Analysis}}  & 16bits & 32bits & 64bits  \\
\hline
\hline
\multirow{4}{*}{CIFAR-10} & $L_{quan}$+ $L_{sim}$ & 0.879 & 0.901 & 0.915 \\
\cline{2-5}
~ & +$L_{kl}$    & 0.911 & 0.920 & 0.928 \\
~ & +$L_{rec}$    & 0.930 & 0.943 & 0.955 \\
\cline{2-5}
~ & \textbf{IPHash}   & \textbf{0.942} & \textbf{0.951} & \textbf{0.958} \\
\hline
\multirow{4}{*}{MSCOCO} & $L_{quan}$+ $L_{sim}$ & 0.647 & 0.756 & 0.836 \\
\cline{2-5}
~ & +$L_{kl}$    & 0.790 & 0.853 & 0.877 \\
~ & +$L_{rec}$    & 0.803 & 0.871 & 0.885 \\
\cline{2-5}
~ & \textbf{IPHash}   & \textbf{0.826} & \textbf{0.875} & \textbf{0.894} \\
\hline
\end{tabular}
\end{table}

\begin{figure}
\subfigure{
\begin{minipage}[t]{0.45\linewidth}
\centering
\includegraphics[width=\linewidth]{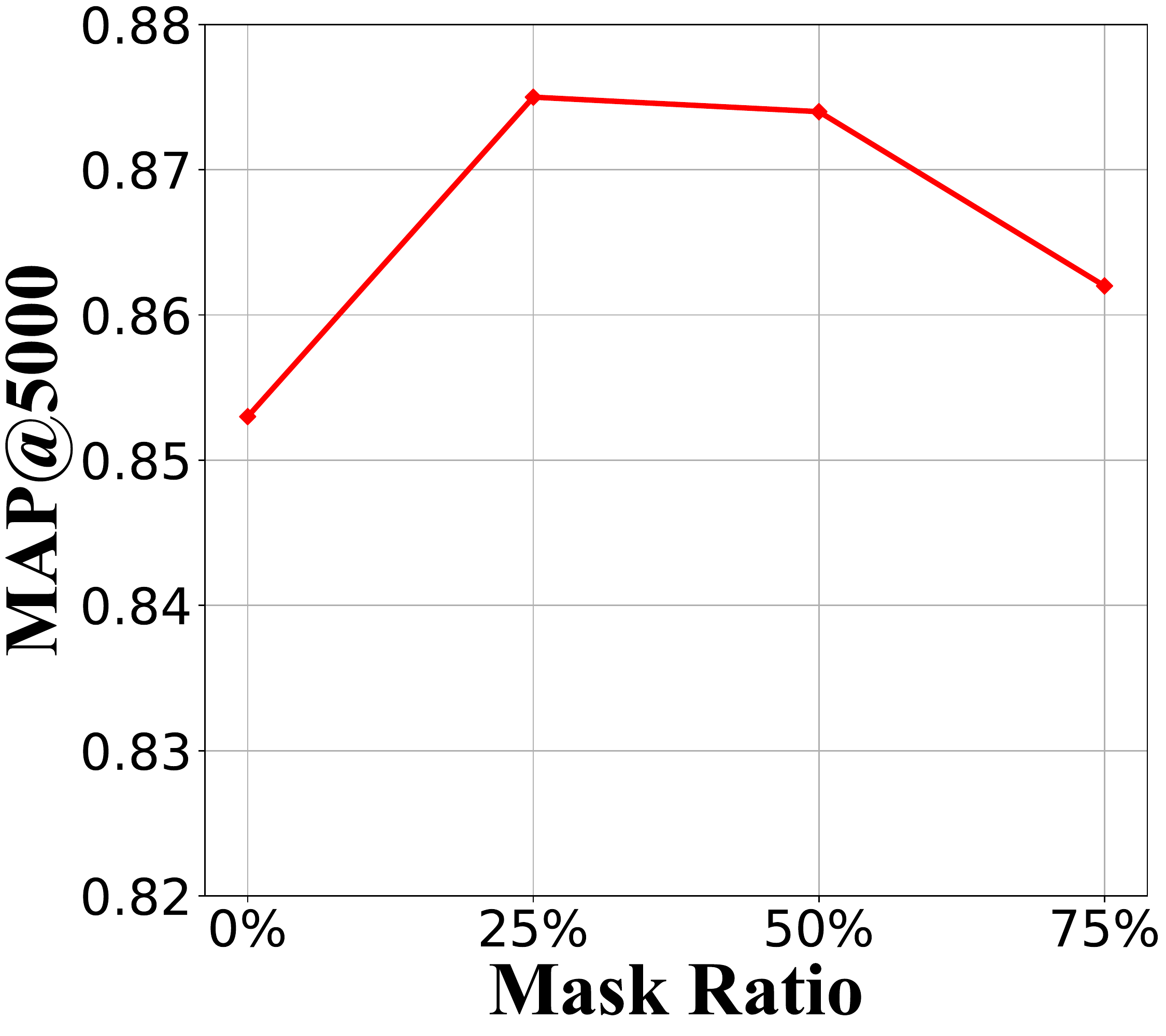}
\end{minipage}
}%
\subfigure{
\begin{minipage}[t]{0.45\linewidth}
\centering
\includegraphics[width=\linewidth]{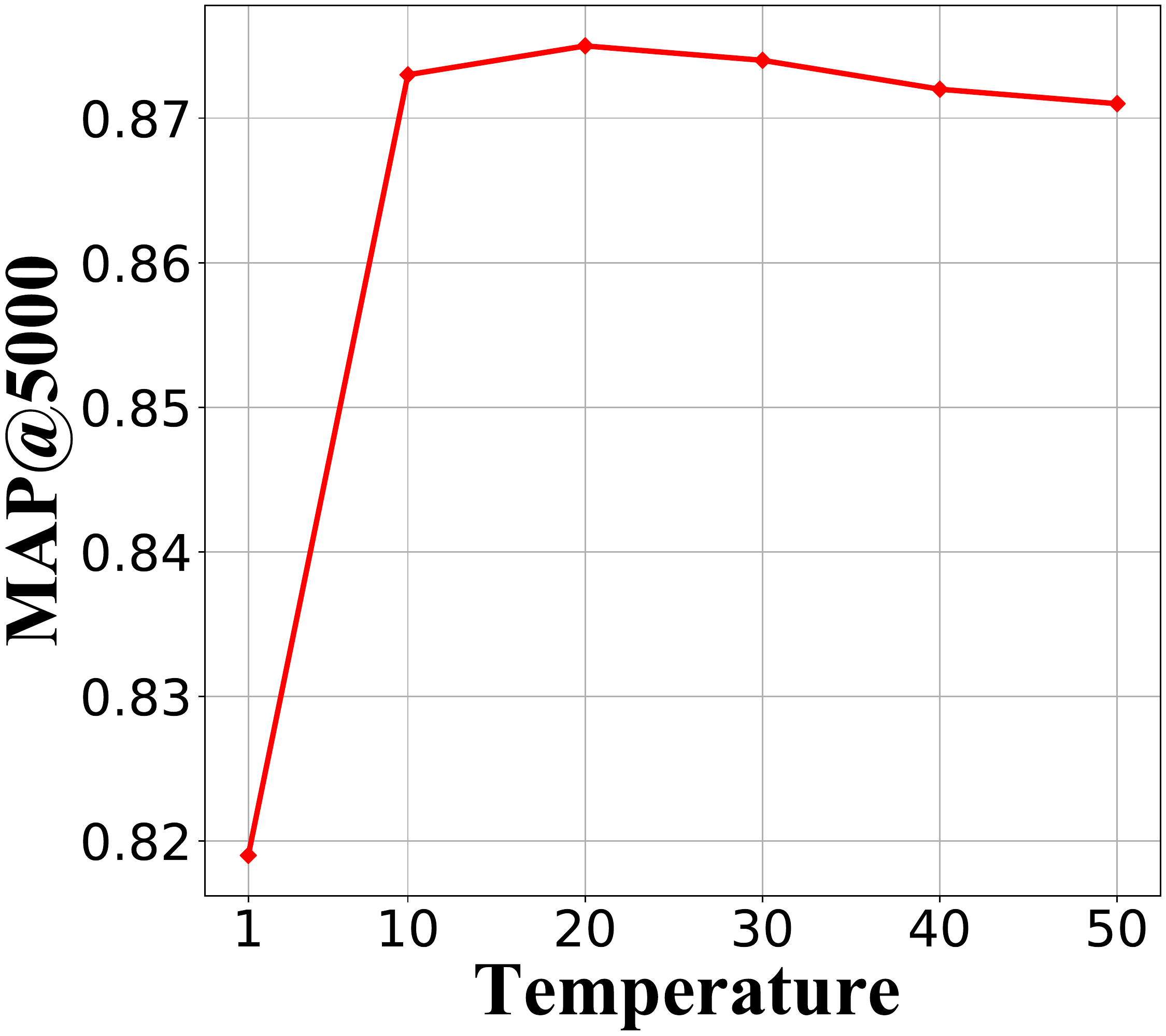}
\end{minipage}
}%
\centering
\caption{Parameter analysis for the temperature and mask ratio with 32-bit hashing codes on MSCOCO.}
\label{parameter_analysis}
\end{figure}

\subsection{Ablation study}

\paragraph{Component Analysis.} To explore the effect of our proposed feature-preserving module and hashing-preserving module, we perform the component analysis on CIFAR-10 and MSCOCO as shown in Table \ref{ablation_study}. Firstly, since the similarity loss $L_{sim}$ and quantization loss $L_{quan}$ have been proposed in the previous methods, e.g, GreedyHash, the baseline method is $L_{sim} + L_{quan}$. To explore the feature-preserving module, we add another baseline $+L_{rec}$ which means that we use the similarity loss, quantization loss and the reconstruct loss. Similar to that, the third baseline is  $+L_{rec}$ which uses the similarity loss, quantization loss, and Kullback-Leibler divergence loss to explore the hashing-preserving module. Finally, the IPHash is the proposed method with the four losses. 


As expected, the performance of adding both modules individually is higher than the baseline, reflecting that the feature-preserving module and hashing-preserving module both have a positive effect on the overall model. In addition, the feature-preserving module brings more improvement since it adjusts the entire ViT module, while the hashing-preserving module only adjusts a fully connected layer. IPHash that combines the two modules has the best performance and the improvement is most obvious at 16-bit, which reveals that our model can effectively reduce the information loss in scenarios with high compression rate.

\paragraph{Parameter Analysis.} To see how the hyper-parameters mask ratio $m$ and temperature $\tau$ influence the performance, we evaluate the model under different values of mask ratio and temperature with 32-bit hashing codes on MSCOCO as shown in the Figure~\ref{parameter_analysis}. Firstly, unlike MAE, which requires a high mask ratio (\textit{e.g.}, 75\%), our IPHash reaches the highest level when the mask ratio is in the range of 25\% to 50\%, because the task of reconstructing the original image of MAE is low-level. Our goal is to keep semantic information. If the mask ratio is too high, it will lead to excessive loss of semantic information. 

As indicated by~\cite{hinton2015distilling}, a higher value of temperature $\tau$ produces a softer probability distribution over classes. As shown in Figure \ref{parameter_analysis}, we can see that a relatively high temperature can improve the performance of the model. After the value of temperature $\tau$ is larger than 10, the performance is similar.  



\section{Conclusion}

In this paper, we present IPHash, which is a novel unsupervised information-preserving hashing method based on the large-scale vision pre-training model, ViT. Firstly, to learn the information-preserving feature extractor, we utilize the random mask to effectively reduce the spatial redundancy and maintain the semantic information of the original image.  With the hashing-preserving module, we can further transfer the rich semantic information of the pre-trained ViT model to hash codes. Extensive experiments are conducted on three benchmark datasets. The results demonstrate that our method significantly outperforms the baselines.

In our future work, we will plan to explore the unsupervised contrastive vision-language pre-training models, \textit{e.g.}, CLIP, and prompt-tuning approaches for unsupervised hashing. 

\bibliographystyle{named}
\bibliography{ijcai22}

\end{document}